\documentclass[11pt,a4paper]{article}
\usepackage[hyperref]{naaclhlt2018}
\usepackage{times}
\usepackage{latexsym}

\usepackage{url}

\usepackage{amsmath}
\usepackage{amssymb}
\usepackage{multirow}
\usepackage{booktabs}

\usepackage{graphicx}
\usepackage{epstopdf}

\aclfinalcopy 

\title{Improving Neural Machine Translation with Conditional Sequence Generative Adversarial Nets}

\author{Zhen Yang$^{1,2}$,  Wei Chen$^1$\footnotemark[1]  ,  Feng Wang$^{1,2}$, Bo Xu$^1$ \\
  $^1$Institute of Automation, Chinese Academy of Sciences \\
  $^2$University of Chinese Academy of Sciences \\
  {\tt \{yangzhen2014, wei.chen.media, feng.wang, xubo\}@ia.ac.cn}}

\date{}

\begin{document}
\maketitle
\begin{abstract}
This paper proposes an approach for applying GANs to NMT. We build a conditional sequence generative adversarial net which comprises of two adversarial sub models, a generator and a discriminator. The generator aims to generate sentences which are hard to be discriminated from human-translated sentences ( i.e., the golden target sentences); And the discriminator makes efforts to discriminate the machine-generated sentences from human-translated ones. The two sub models play a mini-max game and achieve the win-win situation when they reach a Nash Equilibrium. Additionally, the static sentence-level BLEU is utilized as the reinforced objective for the generator, which biases the generation towards high BLEU points. During training, both the dynamic discriminator and the static BLEU objective are employed to evaluate the generated sentences and feedback the evaluations to guide the learning of the generator. Experimental results show that the proposed model consistently outperforms the traditional RNNSearch and the newly emerged state-of-the-art Transformer on English-German and Chinese-English translation tasks.
\end{abstract}

\section{Introduction}
Neural machine translation \cite{Kalchbrenner:13,sutskever:14,cho:14b,bahdanau:14} which directly leverages a single neural network to transform the source sentence into the target sentence, has drawn more and more attention in both academia and industry \cite{shen:15,wu2016google,johnson2016google,Gehring2017Convolutional,Vaswani2017Attention}. This end-to-end NMT typically consists of two sub neural networks. The encoder network reads and encodes the source sentence into the context vector representation; and the decoder network generates the target sentence word by word based on the context vector. To dynamically generate a context vector for a target word being generated, the attention mechanism which enables the model to focus on the relevant words in the source-side sentence is usually deployed. Under the encoder-decoder framework, many variants of the model structure, such as convolutional neural network (CNN) and recurrent neural network (RNN) are proposed \cite{bahdanau:14,Gehring2017Convolutional}. Recently, \cite{Gehring2017Convolutional} propose the Transformer, the first sequence transduction model based entirely on attention, achieving state-of-the-art performance on the English-German and English-French translation tasks. Despite its success, the Transformer, similar to traditional NMT models, is still optimized to maximize the likelihood estimation of the ground word (MLE) at each time step. Such an objective poses a hidden danger to NMT models. That is, the model may generate the best candidate word for the current time step yet a bad component of the whole sentence in the long run. Minimum risk training (MRT) \cite{shen:15} is proposed to alleviate such a limitation by adopting the sequence level objective, i.e., the sentence-level BLEU, for traditional NMT models. Yet somewhat improved, this objective still does not guarantee the translation results to be natural and sufficient. Since the BLEU point is computed as the geometric mean of the modified n-gram precisions \cite{papineni2002bleu:02}, almost all of the existing objectives essentially train NMT models to generate sentences with n-gram precisions as high as possible (MLE can be viewed to generate sentences with high 1-gram precisions). While n-gram precisions largely tell the good sentence apart from the bad one, it is widely acknowledged that higher n-gram precisions do not guarantee better sentences \cite{callison2006re:06,chatterjee2007some}. Additionally, the manually defined objective, i.e., the n-gram precision, is unable to cover all crucial aspects of the data distribution and NMT models may be trained to generate suboptimal sentences \cite{luc2016semantic}\footnote{This paper has been accepted by NAACL-HLT2018.}.

In this paper, to address the limitation mentioned above, we borrow the idea of generative adversarial training from computer vision \cite{goodfellow2014generative,denton2015deep} to directly train the NMT model generating sentences which are hard to be discriminated from human translations. The motivation behind is that while we can not manually define the data distribution of golden sentences comprehensively, we are able to utilize a discriminative network to learn automatically what the golden sentences look like. Following this motivation, we build a conditional sequence generative adversarial net where we jointly train two sub adversarial models: A generator generates the target-language sentence based on the input source-language sentence; And a discriminator, conditioned on the source-language sentence, predicts the probability of the target-language sentence being a human-generated one. During the training process, the generator aims to fool the discriminator into believing that its output is a human-generated sentence, and the discriminator makes efforts not to be fooled by improving its ability to distinguish the machine-generated sentence from the human-generated one. This kind of adversarial training achieves a win-win situation when the generator and discriminator reach a Nash Equilibrium \cite{zhao2016energy,Arora2017Generalization,guimaraes2017objective}. Besides generating the desired distribution, we also want to directly guide the generator with a static and specific objective, such as generating sentences with high BLEU points. To this end, the smoothed sentence-level BLEU \cite{Nakov2012Optimizing} is utilized as the reinforced objective for the generator. During training, we employ both the dynamic discriminator and the static BLEU objective to evaluate the generated sentences and feedback the evaluations to guide the learning of the generator. In summary, we mainly make the following contributions:
\begin{itemize}
\item To the best of our knowledge, this work is among the first endeavors to introduce the generative adversarial training into NMT. We directly train the NMT model to generate sentences which are hard to be discriminated from human translations. The proposed model can be applied to any end-to-end NMT systems.
\item We conduct extensive experiments on English-German and Chinese-English translation tasks and we test two different NMT models, the traditional RNNSearch \cite{bahdanau:14} and the state-of-the-art Transformer. Experimental results show that the proposed approach consistently achieves great success.
\item Last but not least, we propose the smoothed sentence-level BLEU as the static and specific objective for the generator which biases the generation towards achieving high BLEU points. We show that the proposed approach is a weighted combination of the naive GAN and MRT.
\end{itemize}

\section{Background and Related Work}
\subsection{RNNSearch and Transformer}
The RNNSearch is the traditional NMT model which has been widely explored. We follow the de facto standard implementation by \cite{bahdanau:14}. The encoder is a bidirectional gated recurrent units that encodes the input sequence $x=(x_1, \ldots, x_m)$ and calculates the forward sequence of hidden states $(\overrightarrow{h_1}, \ldots, \overrightarrow{h_m})$, and a backward sequence of hidden states $(\overleftarrow{h_1}, \ldots, \overleftarrow{h_m})$. The final annotation vector $h_j$ is calculated by concatenating $\overrightarrow{h_j}$ and $\overleftarrow{h_j}$. The decoder is a recurrent neural network that predicts a target sequence $y=(y_1,\ldots,y_n)$. Each word $y_i$ is predicted on a recurrent hidden state $s_i$, the previously predicted word $y_{i-1}$ and a context vector $c_i$. The $c_i$ is computed as a weighted sum of the encoded annotations $h_j$. The weight $a_{ij}$ of each annotation $h_j$ is computed by the attention mechanism, which models the alignment between $y_i$ and $x_j$.

The Transformer, recently proposed by \cite{Vaswani2017Attention}, achieves state-of-the-art results on both WMT2014 English-German and WMT2014 English-French translation tasks. The encoder of Transformer is composed of a stack of six identical layers. Each layer consists of a multi-head self-attention and a simple position-wise fully connected feed-forward network. The decoder is also composed of a stack of six identical layers. In addition to the two sub-layers in each encoder layer, the decoder inserts a third sub-layer, which performs multi-head attention over the output of the encoder stack. The Transformer can be trained significantly faster than architectures based on recurrent or convolutional layers since it allows for significantly more parallelization.

\subsection{Generative adversarial nets}
Generative adversarial network, has enjoyed great success in computer vision and has been widely applied to image generation \cite{zhu2017unpaired,radford2015unsupervised}. The conditional generative adversarial nets \cite{gauthier2014conditional} apply an extension of generative adversarial network to a conditional setting, which enables the networks to condition on some arbitrary external data. Some recent works have begun to apply the generative adversarial training into the NLP area: \cite{chen2016adversarial} apply the idea of generative adversarial training to sentiment analysis and \cite{zhang2017aspect} use the idea to domain adaptation tasks. For sequence generation problem, \cite{Yu2016SeqGAN} leverage policy gradient reinforcement learning to back-propagate the reward from the discriminator, showing presentable results for poem generation, speech language generation and music generation. Similarly, \cite{zhanggenerating2016} generate the text from random noise via adversarial training. A striking difference from the works mentioned above is that, our work is in the conditional setting where the target-language sentence is generated conditioned on the source-language one. In parallel to our work, \cite{li2017adversarial} propose a similar conditional sequence generative adversarial training for dialogue generation. They use a hierarchical long-short term memory (LSTM) architecture for the discriminator. In contrast to their approach, we apply the CNN-based discriminator for the machine translation task. Furthermore, we propose to utilize the sentence-level BLEU as the specific objective for the generator. Detailed training strategies for the proposed model and extensive quantitative results are reported. We noticed that \cite{Wu2017Adversarial} is exploring the potential of GAN in NMT too. There are some differences in training strategies and experimental settings between \cite{Wu2017Adversarial} and this work. And the most significant difference is that we propose a novel BLEU-reinforced GAN for NMT.

\section{The Approach}
\subsection{Model overview}
\begin{figure}
  \begin{center}
	\includegraphics[width=7.5cm]{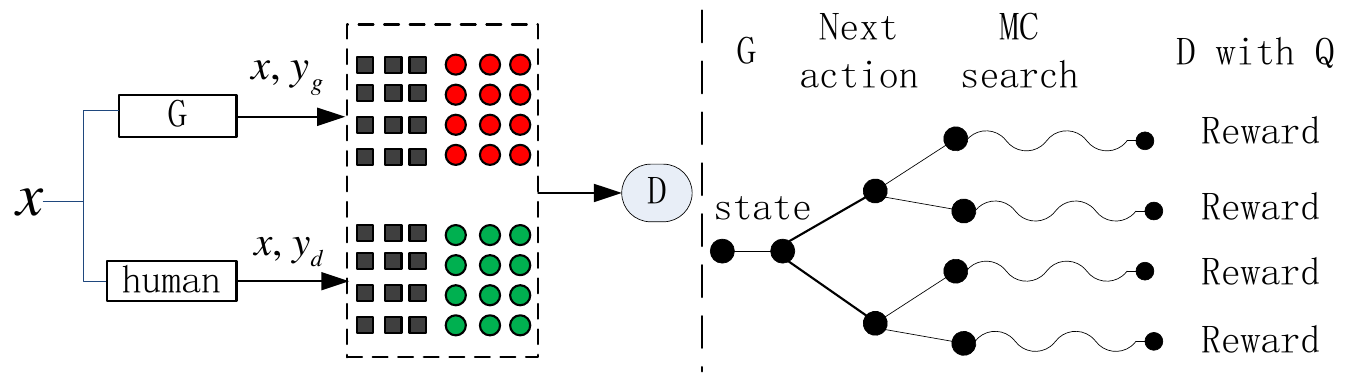}
  \end{center}
  \caption{The Illustration of the proposed BR-CSGAN. Left: D is trained over the real sentence pairs translated by the human and the generated sentence pairs by G. Note that D is a conditional discriminator. Right: G is trained by police gradient where the final reward is provided by D and Q.}
  \label{fig:model_architecture}	
\end{figure}

In this section, we describe the architecture of the proposed BLEU reinforced conditional sequence generative adversarial net (referred to as BR-CSGAN) in detail. The sentence generation process is viewed as a sequence of actions that are taken according to a policy regulated by the generator. In this work, we take the policy gradient training strategies following \cite{Yu2016SeqGAN}. The whole architecture of the proposed model is depicted in figure \ref{fig:model_architecture}. The model mainly consists of three sub modules:

\textbf{Generator} Based on the source-language sentences, the generator \emph{G} aims to generate target-language sentences indistinguishable from human translations.

\textbf{Discriminator} The discriminator \emph{D}, conditioned on the source-language sentences, tries to distinguish the machine-generated sentences from human translations. \emph{D} can be viewed as a dynamic objective since it is updated synchronously with G.

\textbf{BLEU objective} The sentence-level BLEU \emph{Q} serves as the reinforced objective, guiding the generation towards high BLEU points. \emph{Q} is a static function which will not be updated during training.

\subsection{Generator}
Resembling NMT models, the generator \emph{G} defines the policy that generates the target sentence $y$ given the source sentence $x$. The generator takes exactly the same architecture with NMT models. Note that we do not assume the specific architecture of the generator. To verify the effectiveness of the proposed method, we take two different architectures for the generator, the RNNSearch \footnote{https://github.com/nyu-dl/dl4mt-tutorial} and Transformer \footnote{https://github.com/tensorflow/tensor2tensor}.


\subsection{Discriminator}
Recently, the deep discriminative models such as the CNN and RNN have shown a high performance in complicated sequence classification tasks. Here, the discriminator is implemented based on the CNN architecture.

Since sentences generated by the generator have variable lengths, the CNN padding is used to transform the sentences to sequences with fixed length $T$, which is the maximum length set for the output of the generator. Given the source-language sequence $x_1, \ldots, x_T$ and target-language sequence $y_1, \ldots, y_T$, we build the source matrix $X_{1:T}$ and target matrix $Y_{1:T}$ respectively as:
\begin{equation}X_{1:T}=x_1;x_2;\ldots;x_T\end{equation}
and
\begin{equation}Y_{1:T}=y_1;y_2;\ldots;y_T\end{equation}
where $x_t$, $y_t \in R^k$ is the $k$-dimensional word embedding and the semicolon is the concatenation operator. For the source matrix $X_{1:T}$, a kernel ${w_j}\in R^{l \times k}$ applies a convolutional operation to a window size of $l$ words to produce a series of feature maps:
\begin{equation}\label{equa:BN}c_{ji}=\rho(BN(w_j \otimes X_{i:i+l-1}+b))\end{equation}
where $\otimes$ operator is the summation of element-wise production and $b$ is a bias term. $\rho$ is a non-linear activation function which is implemented as ReLu in this paper. Note that the batch normalization \cite{ioffe2015batch} which accelerates the training significantly, is applied to the input of the activation function ($BN$ in equation \ref{equa:BN}). To get the final feature with respect to kernel $w_j$, a max-over-time pooling operation is leveraged over the feature maps:
\begin{equation}\widetilde{c}_j=max\{c_{j1},\ldots,c_{j{T-l+1}}\}\end{equation}
We use various numbers of kernels with different window sizes to extract different features, which are then concatenated to form the source-language sentence representation $c_x$. Identically, the target-language sentence representation $c_y$ can be extracted from the target matrix $Y_{1:T}$. Finally, given the source-language sentence, the probability that the target-language sentence is being real can be computed as:
\begin{equation} \label{probability} p=\sigma(V[c_x;c_y])\end{equation}
where $V$ is the transform matrix which transforms the concatenation of $c_x$ and $c_y$ into a 2-dimension embedding and $\sigma$ is the logistic function.

\subsection{BLEU objective}
We apply the smoothed sentence-level BLEU as the specific objective for the generator. Given the generated sentence $y_g$ and the the ground true sentence $y_d$, the objective $Q$ calculates a reward $Q(y_g,y_d)$, which measures the n-gram precisions of the generated sentence $y_g$. Identical to the output of the discriminator, the $Q(y_g, y_d)$ also ranges from zero to one, which makes it easier to fuse $Q$ and $D$.

\subsection{Policy Gradient Training}
Following \cite{Yu2016SeqGAN}, the objective of the generator $G$ is defined as to generate a sequence from the start state to maximize its expected end reward. Formally, the objective function is computed as:
\begin{equation}
\resizebox{0.95\hsize}{!}{$
    \label{eq:obj}
	J(\theta)= \sum\limits_{Y_{1:T}} G_{\theta}(Y_{1:T}|X)\cdot R_{D,Q}^{G_{\theta}}(Y_{1:T-1},X, y_T, Y^*)$} \nonumber
\end{equation}
where $\theta$ represents the parameters in $G$, $Y_{1:T}=y_1,\ldots,y_{T}$ indicates the generated target sequence, $X$ is the source-language sentence, $Y^*$ represents the ground true target sentence.
$R_{D,Q}^{G_\theta}$ is the action-value function of a target-language sentence given the source sentence $X$, i.e. the expected accumulative reward starting from the state $(Y_{1:T-1},X)$, taking action $y_T$, and following the policy $G_{\theta}$. To estimate the action-value function, we consider the estimated probability of being real by the discriminator $D$ and the output of the BLEU objective $Q$ as the reward:
\begin{equation}
\resizebox{0.95\hsize}{!}{$
\begin{aligned}
	&R_{D,Q}^{G_\theta}(Y_{1:T-1},X, y_T, Y^*)=\\
   & \lambda(D(X, Y_{1:T})- b(X,Y_{1:T}))+(1-\lambda)Q(Y_{1:T}, Y^*) \nonumber
\end{aligned}$}
\end{equation}
where b(X,Y) denotes the baseline value to reduce the variance of the reward. Practically, we take b(X,Y) as a constant, 0.5 for simplicity. And the $\lambda$ is a hyper-parameter. The question is that, given the source sequence, $D$ only provides a reward value for a finished target sequence. If $Y_{1:T}$ is not a finished target sequence, the value of $D(X,Y_{1:T})$ makes no sense. Therefore, we cannot get the action-value for an intermediate state directly. To evaluate the action-value for an intermediate state, the Monte Carlo search under the policy of $G_{\theta}$ is applied to sample the unknown tokens. Each search ends until the end of sentence token is sampled or the sampled sentence reaches the maximum length. To obtain more stable reward and reduce the variance, we represent an N-time Monte Carlo search as:
\begin{equation}
\resizebox{.80\hsize}{!}{$
	\{Y_{1:T_{1}}^1,\ldots,Y_{1:T_N}^N\}=MC^{G_{\theta}} ((Y_{1:t},X),N)$} \nonumber
\end{equation}
where $T_i$ represents the length of the sentence sampled by the $i$'th Monte Carlo search. $(Y_{1:t},X)=(y_1,\ldots,y_t,X)$ is the current state and $Y_{t+1:T_N}^N$ is sampled based on the policy $G_\theta$. The discriminator provides $N$ rewards for the sampled $N$ sentences respectively. The final reward for the intermediate state is calculated as the average of the $N$ rewards. Hence, for the target sentence with the length $T$, we compute the reward for $y_t$ in the sentence level as:
\begin{equation}
\resizebox{0.99\hsize}{!}{$
\begin{aligned}
& R_{D,Q}^{G_{\theta}}(Y_{1:t-1},X, y_T, Y^*) = \label{eq:q-mcs} \\
&\left\{
\begin{array}{lcl}
\frac{1}{N}\sum\limits_{n=1}^N \lambda (D(X, Y_{1:T_n}^n)-b(X,Y_{1:T_n}^n)) +(1-\lambda)Q(Y_{1:T_n}, Y*)&
\text{$t<T$} \\
\lambda (D(X,Y_{1:t})-b(X,Y_{1:t}))+(1-\lambda)Q(Y_{1:t}, Y^*) & \text{$t=T$}
\end{array}
\right. \nonumber
\end{aligned}$}
\end{equation}
Using the discriminator as a reward function can further improve the generator iteratively by dynamically updating the discriminator. Once we get more realistic generated sequences, we re-train the discriminator as:
\begin{equation}
\resizebox{0.99\hsize}{!}{$\min-\mathbb{E}_{X,Y \in P_{data}}[\log D(X,Y)]-\mathbb{E}_{X,Y \in G}[\log(1-D(X,Y))]$}\nonumber
\end{equation}
After updating the discriminator, we are ready to re-train the generator. The gradient of the objective function $J(\theta)$ w.r.t the generator's parameter $\theta$ is calculated as:
\begin{equation}
\resizebox{0.99\hsize}{!}{$
\begin{array}{lcl}
\nabla J(\theta)=
\frac{1}{T}\sum\limits_{t=1}^T\sum\limits_{y_t}R_{D,Q}^{G_{\theta}}(Y_{1:t-1},X, y_T, Y^*)\cdot\nabla_{\theta}(G_{\theta}(y_t|Y_{1:t-1},X)) &\\
=\frac{1}{T}\sum\limits_{t=1}^T\mathbb{E}_{y_t \in G_{\theta}} [R_{D,Q}^{G_{\theta}}(Y_{1:t-1},X, y_T, Y^*)\cdot\nabla_{\theta}\log p(y_t|Y_{1:t-1},X)]
\end{array}\nonumber$}
\end{equation}

\begin{table*}[htb]
\centering
\scalebox{0.95}{
\begin{tabular}{c|cccc|c}
\toprule[2pt]
\multirow{2}*{Model} &	\multicolumn{4}{c|}{Chinese-English} &  English-German \\

                     & NIST03 & NIST04 & NIST05 & average & newstest2014 \\
\midrule[1pt]
RNNSearch \cite{bahdanau:14} &33.93 &35.67 &32.24 &33.94 &21.2 \\
Transformer \cite{Vaswani2017Attention} &42.23 &42.17 &41.02 &41.80 &27.30\\
\hline
\hline
RNNSearch+BR-CSGAN($\lambda=1.0$) &35.21 &36.51 &33.45 &35.05 &22.1\\
RNNSearch+BR-CSGAN($\lambda=0.7$) &35.97 &37.32 &34.03 & \textbf{35.77}&\textbf{22.89}\\
RNNSearch+BR-CSGAN($\lambda=0$) &34.57 &35.93 &33.07 &34.52 &21.75\\
\hline
Transformer+BR-CSGAN($\lambda=1.0$) &42.67 &42.79 &41.54 &42.30 &27.75\\
Transformer+BR-CSGAN($\lambda=0.8$) &43.01 &42.96 &41.86 &\textbf{42.61} &\textbf{27.92}\\
Transformer+BR-CSGAN($\lambda=0$) &42.41 &42.74 &41.29 &42.14 &27.49\\
\bottomrule[2pt]
\end{tabular}}
\caption{BLEU score on Chinese-English and English-German translation tasks. The hyper-parameter $\lambda$ is selected according to the development set. For the Transformer, following \cite{Vaswani2017Attention}, we report the result of a single model obtained by averaging the 5 checkpoints around the best model selected on the development set.}
\label{tab:main_result}
\end{table*}

\subsection{Training strategies}
GANs are widely criticized for its unstable training since the generator and discriminator need to be carefully synchronized. To make this work easier to reproduce, this paper gives detailed strategies for training the proposed model.
		
Firstly, we use the maximum likelihood estimation to pre-train the generator on the parallel training set until the best translation performance is achieved. Then, generate the machine-generated sentences by using the generator to decode the training data. We simply use the greedy sampling method instead of the beam search method for decoding. Next, pre-train the discriminator on the combination of the true parallel data and the machine-generated data until the classification accuracy achieves at $\xi$. Finally, we jointly train the generator and discriminator. The generator is trained with the policy gradient training method. However, in our practice, we find that updating the generator only with the simple policy gradient training leads to unstableness. To alleviate this issue, we adopt the teacher forcing approach which is similar to \cite{Lamb2016Professor,li2017adversarial}. We directly make the discriminator to automatically assign a reward of 1 to the golden target-language sentence and the generator uses this reward to update itself on the true parallel example. We run the teacher forcing training for one time once the generator is updated by the policy gradient training. After the generator gets updated, we use the new stronger generator to generate $\eta$ more realistic sentences, which are then used to train the discriminator. Following \cite{arjovsky2017wasserstein}, we clamp the weights of the discriminator to a fixed box ( [$-\epsilon$,$\epsilon$] ) after each gradient update. We perform one optimization step for the discriminator for each step of the generator. In our practice, we set $\xi$ as 0.82, $\eta$ as 5000, $\epsilon$ as 1.0 and the $N$ for Monte Carlo search as 20.

\section{Experiments and Results}
We evaluate our BR-CSGAN on English-German and Chinese-English translation tasks and we test two different architectures for the generator, the traditional RNNSearch and the newly emerged state-of-the-art Transformer.

\subsection{Data sets and preprocessing}
English-German: For English-German translation, we conduct our experiments on the publicly available corpora used extensively as benchmark for NMT systems, WMT'14 En-De. This data set contains 4.5M sentence pairs \footnote{http://nlp.stanford.edu/projects/nmt}. Sentences are encoded using byte-pair encoding \cite{Sennrich2015Neural}, which has a shared source-target vocabulary of about 37000 tokens.
We report results on newstest2014. The newstest2013 is used as validation.

Chinese-English: For Chinese-English translation, our training data consists of 1.6M sentence pairs randomly extracted from LDC corpora \footnote{LDC2002L27, LDC2002T01, LDC2002E18, LDC2003E07, LDC2004T08, LDC2004E12, LDC2005T10}. Both the source and target sentences are encoded with byte-pair encoding and the tokens in the source and target vocabulary is about 38000 and 34000 respectively \footnote{When doing BPE for Chinese, we need to do word segmentation first and the following steps are the same with BPE for English.}. We choose the NIST02 as the development set. For testing, we use NIST03, NIST04 and NIST05 data sets.

To speed up the training procedure, sentences of length over 50 words are removed when we conduct experiments on the RNNSearch model. This is widely used by previous works \cite{ranzato:15,shen:15,yang-EtAl:2016:COLING}.

\subsection{Model parameters and evaluation}
For the Transformer, following the base model in \cite{Vaswani2017Attention}, we set the dimension of word embedding as 512, dropout rate as 0.1 and the head number as 8. The encoder and decoder both have a stack of 6 layers. We use beam search with a beam size of 4 and length penalty $\alpha=0.6$. For the RNNSearch, following \cite{bahdanau:14}, We set the hidden units for both encoders and decoders as 512. The dimension of the word embedding is also set as 512. We do not apply dropout for training the RNNSearch. During testing, we use beam search with a beam size of 10 and length penalty is not applied.

All models are implemented in TensorFlow \cite{tensorflow2015-whitepaper} and trained on up to four K80 GPUs synchronously in a multi-GPU setup on a single machine \footnote{The code we used to train and evaluate our models can be found at https://github.com/ZhenYangIACAS/NMT\_GAN}. We stop training when the model achieves no improvement for the tenth evaluation on the development set. BLEU \cite{papineni2002bleu:02} is utilized as the evaluation metric. We apply the script \emph{mteval-v11b.pl} to evaluate the Chinese-English translation and utilize the script \emph{multi-belu.pl} for English-German translation \footnote{https://github.com/moses-smt/mosesdecoder/blob/617e8c8/scripts/generic/{multi-bleu.perl;mteval-v11b.pl}}.

\subsection{Main results}
The model of RNNSearch is optimized with the mini-batch of 64 examples. It takes about 30 hours to pre-train the RNNSearch on the Chinese-English data set and 46 hours on the English-German data set. During generative adversarial training, it takes about 35 hours on the Chinese-English data set and about 50 hours on the English-German data set. For the Transformer, each training batch contains a set of sentence pairs containing approximately 25000 source tokens and 25000 target tokens. On the Chinese-English data set, it takes about 15 hours to do pre-training and 20 hours to do generative adversarial training. On the English-German data set, it takes about 35 hours for the pre-training and 40 hours for the generative adversarial training.

Table \ref{tab:main_result} shows the BLEU score on Chinese-English and English-German test sets. On the RNNSearch model, the naive GAN (i.e., the line of RNNSearch+BR-CSGAN ($\lambda$=1) in table \ref{tab:main_result}) achieves improvement up to +1.11 BLEU points averagely on Chinese-English test sets and +0.9 BLEU points on English-German test set. Armed with the BLEU objective, the BR-CSGAN (the line of RNNSearch+BR-CSGAN ($\lambda$=0.7)) leads to more significant improvements, +1.83 BLEU points averagely on Chinese-English translation and +1.69 BLEU points on English-German translation. We also test the translation performance when the RNNSearch is only guided by the static BLEU objective (the line of RNNSearch+BR-CSGAN ($\lambda$=0)), and we only get +0.58 BLEU points improvement on Chinese-English translation and +0.55 BLEU points improvement on English-German. Experiments on the Transformer show the same trends. While the Transformer has achieved state-of-the-art translation performances, our approach still achieves +0.81 BLEU points improvement on Chinese-English translation and +0.62 BLEU points improvement on English-German.

These results indicate that the proposed BR-CSGAN consistently outperforms the baselines and it shows better translation performance than the naive GAN and the model guided only by the BLEU objective.

\section{Analysis}
\subsection{Compared with MRT}
We show that MRT \cite{shen:15} is an extreme case of our approach. Considering a sentence pair $(x,y)$, the training objective of MRT is calculated as
\begin{equation}
\widehat{J}(\theta')=\sum\limits_{y^s \in S(x)}p(y^s|x;\theta')\Delta(y^s,y) \nonumber
\end{equation}
where $\Delta(y^s,y)$ is a loss function (i.e., the sentence-level BLEU used in this paper) that measures the discrepancy between a predicted translation $y^s$ and the training example $y$, $S(x)$ represents the set which contains all of the predictions given the input $x$, and $\theta'$ is the parameters of the NMT model. Unfortunately, this objective is usually intractable due to the exponential search space. To alleviate this problem, a subset of the search space is sampled to approximate this objective. In this paper, when we set $\lambda$ as zero, the objective for the proposed BR-CSGAN comes to
\begin{equation}
    \label{eq:obj_zero}
	J(\theta)_{\lambda=0}= \sum\limits_{Y_{1:T}} G_{\theta}(Y_{1:T}|X)\cdot Q(Y_{1:T}, Y^*) \nonumber
\end{equation}
where the $Q(Y_{1:T}, Y^*)$ is also a loss function between the predicted translation $Y_{1:T}$ and the training example $Y^*$. It is easy to be found that, under this condition (i.e., $\lambda$ set as zero), the proposed BR-CSGAN optimizes almost the same objective with MRT. The only difference is that the reinforcement learning procedure is utilized in BR-CSGAN to maximize the total reward and MRT instead applies random sampling to approximate the risk. Actually, the BR-CSGAN is a weighted sum of the naive GAN ($\lambda$=1) and MRT ($\lambda$=0), and it incorporates the advantages of the two approaches. Specifically, compared to naive GAN which is trained without specific objective guidance, BR-CSGAN utilizes the BLEU objective to guide the generator to generate sentences with higher BLEU points. And compared to MRT which is trained only with the static objective, the BR-CSGAN applies a dynamic discriminator which updates synchronously with the generator, to feedback the dynamic rewards for the generator. Table \ref{tab:cpm-MRT_CSGAN} compares the translation performance between the MRT and BR-CSGAN on Chinese-English and English-German translation tasks. We only conduct experiments on the RNNSearch because we only get the open-source implementation of MRT on the RNNSearch \footnote{The open-source implementation can be found at: https://github.com/EdinburghNLP/nematus}. Results show that the proposed BR-CSGAN consistently outperforms the MRT on the Chinese-English and English-German translations.

\begin{table}[htb]
\centering
\scalebox{0.75}{
\begin{tabular}{c|c|c}
\toprule[2pt]
\multirow{2}*{Model} &	Chinese-English &  English-German \\
        & average & newstest2014 \\
\midrule[1pt]
RNNSearch & 33.94 & 21.2 \\
MRT \cite{shen:15} &34.64 &21.6 \\
BR-CSGAN($\lambda=0.7$) &35.77 &22.89\\
\bottomrule[2pt]
\end{tabular}}
\caption{BLEU score on Chinese-English and English-German translation tasks for MRT and BR-CSGAN.}
\label{tab:cpm-MRT_CSGAN}
\end{table}

\subsection{When to stop pre-training}
\label{sub:init_acc}
\begin{figure}
   \begin{center}
   \includegraphics[width=7.5cm]{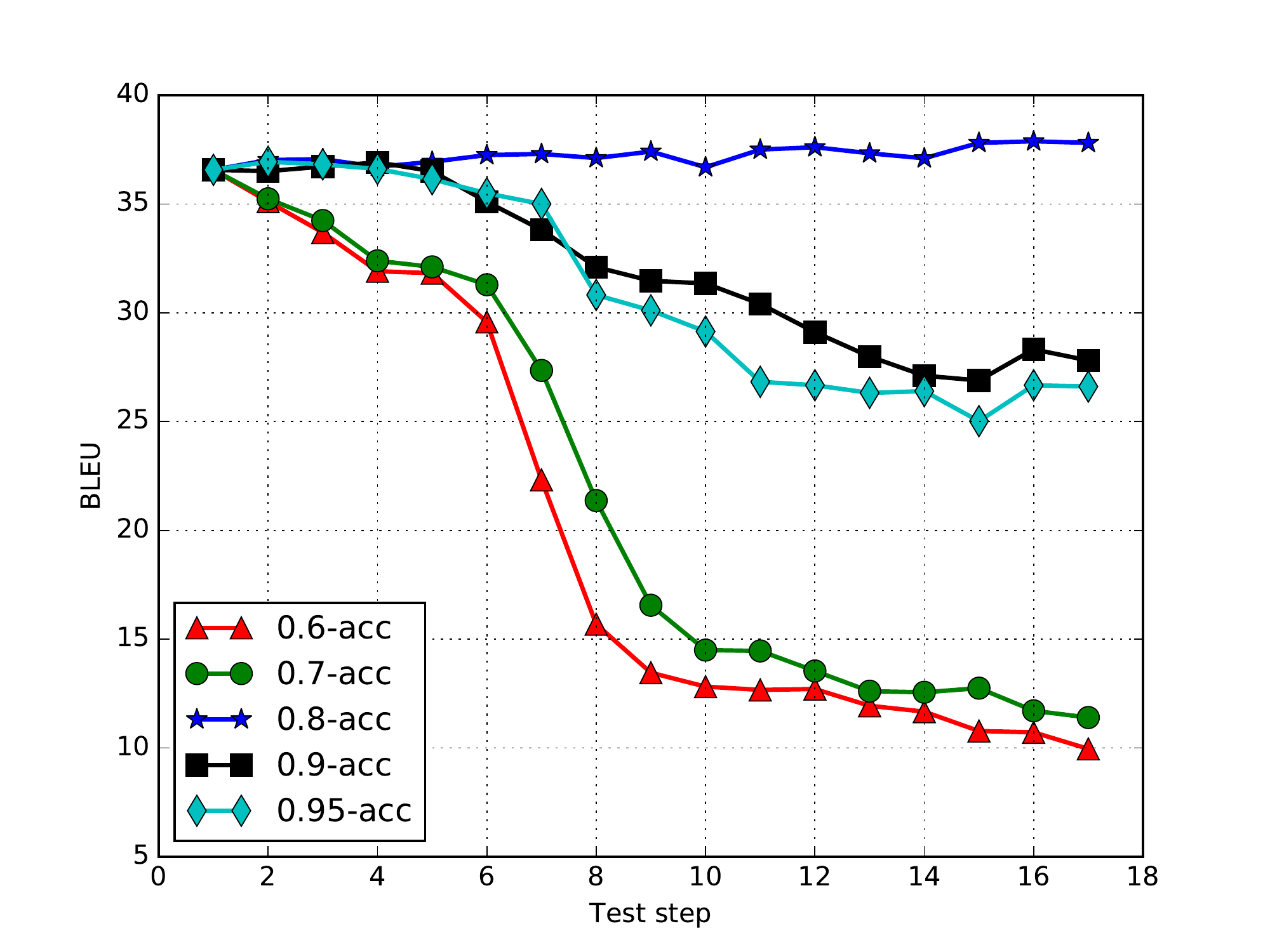}
   \end{center}
   \caption{\label{fig:Initial-accu}BLEU score on the development set for the BR-CSGAN where the discriminators have different initial accuracy. "0.6-acc" means the initial accuracy is 0.6. We report the results on the Chinese-English translation tasks. RNNSearch is taken as the generator.}	
\end{figure}

The initial accuracy $\xi$ of the discriminator which is viewed as a hyper-parameter, can be set carefully during the process of pre-training. A natural question is that when shall we end the pre-training. Do we need to pre-train the discriminator with the highest accuracy?  To answer this question, we test the impact of the initial accuracy of the discriminator. We pre-train five discriminators which have the accuracy as 0.6, 0.7, 0.8, 0.9 and 0.95 respectively. With the five discriminators, we train five different BR-CSGAN models (with the generator as RNNSearch and $\lambda$ set as 0.7) and test their translation performances on the development set at regular intervals. Figure \ref{fig:Initial-accu} reports the results and we can find that the initial accuracy of the discriminator shows great impacts on the translation performance of the proposed BR-CSGAN. From figure \ref{fig:Initial-accu}, we show that the initial accuracy of the discriminator needs to be set carefully and either it is too high (0.9 and 0.95) or too low (0.6 and 0.7), the model performs badly \footnote{To make the illustration simple and clear, we only depict the results when the RNNSearch acts as the generator.}.  This suggests that it is important for the generator and discriminator to keep a balanced relationship at the beginning of the generative adversarial training. If the discriminator is too strong, the generator is always penalized for its bad predictions and gets no idea about right predictions. Hence, the generator is discouraged all the time and the performance gets worse and worse. On the other hand, if the discriminator is too weak, the discriminator is unable to give right guidance for the generator, i.e. the gradient direction for updating the generator is random. Empirically, we pre-train the discriminator until its accuracy reaches around 0.8.

\begin{table}[htb]
			\centering
             \scalebox{0.90}{
				\begin{tabular}{c|c|c|c|c}
					\toprule[2pt]
					$N$ & NIST02	&	NIST03	&	NIST04 &NIST05\\
					\midrule[1pt]
					0 & 36.87 & 33.93 & 35.67 & 32.24 \\
					5 & - &- & - & -\\
					10 & - & - &- & -\\
					15 & 37.34 &34.91 &36.09 & 33.45\\
					20 & 38.58 & 35.97 &37.32 & 34.03\\
					25 & 38.65& 36.04 &37.52 & 33.91\\
					30 & 38.74 &36.01 & 37.54 & 33.76\\
					\bottomrule[2pt]
				\end{tabular}}
				\caption{\label{tab:N-MontoCarlo} The translation performance of the BR-CSGAN with different $N$ for Monte Carlo search. "-" means that the proposed model shows no improvement than the pre-trained generator or it can not be trained stably. With $N$ set as 0, it is referred to as the pre-trained generator. Similarly, we only report results on the RNNSearch and $\lambda$ is set as 0.7.}
\end{table}

\subsection{Sample times for Monto Carol search}
We are also curious about how the sample times $N$ for Monte Carlo search affects the translation performance. Intuitively, if $N$ is set as a small number, the intermediate reward for each word may be incorrect since there is a large variance for the Monto Carol search when the sample time is too small. And if otherwise, the computation shall be very time consuming because we need to do much more sampling. Therefore, there is a trade-off between the accuracy and computation complexity here. We investigate this problem on the Chinese-English translation task. Table \ref{tab:N-MontoCarlo} presents the translation performance of the BR-CSGAN on the test sets when the $N$ are set from 5 to 30 with interval 5. From table \ref{tab:N-MontoCarlo}, the proposed model achieves no improvement than the baseline (i.e., the pre-trained generator) when $N$ are set less than 15 and the BLEU scores are not reported on the table. As a matter of fact, the translation performance of the model gets worse and worse. We conjecture that the approximated reward is far from the expected reward due to the large variance when $N$ is set too small, and gives wrong gradient directions for model updating. Since the training for GAN is not stable, the wrong gradient direction exacerbates the unstableness and results in the BLEU getting worse and worse. With the increasing of $N$, the translation performance of the model gets improved. However, with $N$ set larger than 20, we get little improvement than the model with $N$ set as 20 and the training time exceeds our expectation.
		
\section{Conclusion and Future Work}
In this work, we propose the BR-CSGAN which leverages the BLEU reinforced generative adversarial net to improve the NMT. We show that the proposed approach is a weighted combination of the naive GAN and MRT. To verify the effectiveness of our approach, we test two different architectures for the generator, the traditional RNNSearch and the state-of-the-art Transformer. Extensive experiments on Chinese-English and English-German translation tasks show that our approach consistently achieves significant improvements. In the future, we would like to try multi-adversarial framework which consists of multi discriminators and generators for GAN.

\section*{Acknowledgements}
This work is supported by the National Key Research and Development Program of China under Grant No. 2017YFB1002102, and Beijing Engineering Research Center under Grant No. Z171100002217015. We would like to thank Xu Shuang for her preparing data used in this work. Additionally, we also want to thank Chen Zhineng, Wang Wenfu and Zhao Yuanyuan for their invaluable discussions on this work.

\bibliography{naaclhlt2018}
\bibliographystyle{acl_natbib}

\end{document}